\documentclass{article}
\usepackage{amsmath,graphicx,mlspconf}
\usepackage{accents}
\copyrightnotice{U.S.\ Government work not protected by U.S.\ copyright}

\copyrightnotice{978-1-7281-6338-3/21/\$31.00 {\copyright}2021 Crown}

\copyrightnotice{978-1-7281-6338-3/21/\$31.00
{\copyright}2021 European Union}

\copyrightnotice{978-1-7281-6338-3/21/\$31.00 {\copyright}2021 IEEE}

\toappear{2021 IEEE International Workshop on Machine Learning for Signal Processing, Oct.\ 25--28, 2021, Gold Coast, Australia}

\title{Random Feature Approximation for Online Nonlinear Graph Topology Identification}

\author{
\IEEEauthorblockA{\textit{Rohan Money, Joshin Krishnan, Baltasar Beferull-Lozano}\\
\textit{WISENET Center, Department of Information and Communication Technologies} \\
\textit{University of Agder,}
Grimstad, Norway \\
\{rohan.t.money; joshin.krishnan; baltasar.beferull\}@uia.no}
}

\name{Rohan Money, Joshin Krishnan, Baltasar Beferull-Lozano\thanks{Estimating the unknown causal dependencies among This work was supported by the IKTPLUSS INDURB grant 270730/O70 and SFI Offshore Mechatronics grant 237896/O30.}}
\address{%
   WISENET Center, Department of Information and Communication Technologies \\
   University of Agder\\
   Grimstad, Norway,\\
}
\usepackage{bbold}
\usepackage{amsmath,amsthm, mathtools, amssymb, amsfonts, amsbsy,fixmath}
\usepackage[capitalise]{cleveref}
\usepackage{algorithmic}
\usepackage{graphicx}
\usepackage{textcomp}
\usepackage{xcolor}
\usepackage[ruled,vlined]{algorithm2e}
\usepackage{subcaption}

   \usepackage{bibspacing}

\def\BibTeX{{\rm B\kern-.05em{\sc i\kern-.025em b}\kern-.08em
    T\kern-.1667em\lower.7ex\hbox{E}\kern-.125emX}}

\def\cbr#1{\left\lbrace #1 \right\rbrace} 
\def\sbr#1{\biggl [ #1 \biggr ]} 
\def\nbr#1{\left( #1\right)}
\def\bs #1{\boldsymbol{#1}}
\def\bm#1{\mathbf{#1}}

\newcommand{\cB}[1]{\textcolor{black}{#1}}

    \def\p {{(p)}}
\def\n {{n'}}
\begin{document}
%\ninept

\maketitle

\begin{abstract}
Online topology estimation of graph-connected time series is challenging, especially since the causal dependencies in many real-world networks are nonlinear. In this paper, we propose a kernel-based algorithm for graph topology estimation. The algorithm uses a Fourier-based Random feature approximation to tackle the curse of dimensionality associated with the kernel representations. Exploiting the fact that the real-world networks often exhibit sparse topologies, we propose a group lasso based optimization framework, which is solve using an iterative composite objective mirror descent method, yielding an online algorithm with fixed computational complexity per iteration. The experiments conducted on real and synthetic data show that the proposed method outperforms its competitors.

\end{abstract}
\begin{keywords}
Nonlinear topology identification, Causality, Directed graphs, Kernels, Random feature approximation.
\end{keywords}
\section{Introduction}
The amount of data generated from interconnected networks such as sensor networks, financial time-series, brain-networks, etc., are increasing rapidly. Extraction of meaningful information from such interconnected data, represented in the form of a graph can have many practical applications such as,  signal denoising  \cite{desmol2021}, change point detection \cite{dynl.m2018}, time  series prediction \cite{onlB.Z2017}, etc. Many of the functional relationships in such networks are causal and identification of this causal graph structure is termed topology identification.
%\cite{equc.j2011}
Many real world causal  systems can be well described using vector autoregressive model (VAR) as naturally most of the dependencies are time-lagged in nature. Moreover under causal sufficiency, VAR  causality implies well known Granger causality \cite{newlut2005}.
%The relationships between these interdependent data can be represented in the form of a graph with certain properties.

Topology identification based on the linear VAR model has been well-studied. In \cite{onlzam2019}, an efficient way to estimate linear VAR coefficients from streaming data is proposed. However, such linear VAR models fail to capture the real-world nonlinear dependencies.A novel nonlinear VAR topology identification is proposed in \cite{nonshe2019} in which, the kernels are used to linearize the nonlinear dependencies by mapping them to a higher-dimensional Hilbert space. However, being a batch-based approach, \cite{nonshe2019} is computationally expensive and is not suitable for identifying the time-varying topologies.

The above shortcomings are tackled by kernel-based online algorithms \cite{only.s2018}, \cite{onlroh2021}. In \cite{onlroh2021}, sparse VAR coefficients are recursively estimated using a composite objective mirror decent (COMID) approach, whereas \cite{only.s2018} uses functional stochastic gradient descent (FSGD), followed by soft-thresholding. However, the kernel-based representations have a major drawback of unaffordable growth of computational complexity and memory requirement, which is commonly known as the ``curse of dimensionality". Both \cite{only.s2018} and \cite{onlroh2021} propose to circumvent this issue by restricting the numeric calculation to a limited number of time-series samples using a time window, which results in suboptimal performance.

A standard procedure to address the curse of dimensionality is to invoke the kernel dictionaries \cite{para.k2017}. Often, the dictionary elements are selected based on a budget maintaining strategy. In large-scale machine learning problems, the dictionary size can go prohibitively high in order to maintain the budget. Recently, the random feature (RF) approximation \cite{ranrah2007} techniques are gaining popularity in approximating the kernels, which are shown to yield promising results compared to the budget maintaining strategies \cite{ranrah2007,larjin2016}. %The work \cite{ranrah2007} proposed a way to approximate shift-invariant kernels in terms of a fixed number of random features, and the method is termed as random feature approximation.}

In this work, we use RF approximation to avoid the curse of dimensionality in learning  nonlinear VAR models. We approximate shift-invariant Gaussian kernels using a fixed number of random Fourier features. The major contributions of this paper are \textbf{i)} formulation of a kernel-based optimization framework in the function space, \textbf{ii)} reformulation of \textbf{i)} to a parametric optimization using RF approximation, and \textbf{iii)} an online  algorithm to estimate the sparse nonlinear VAR coefficients using COMID updates. We provide numerical results showing the proposed method outperforms the state-of-the-art topology identification algorithms.
%This work is based on some key assumptions such as causal sufficiency and sparse and slowly varying dependency. 
\section{Kernel Representation}
%\subsection{Non-linear topology identification}
\cB{Consider a multi-variate time series with $N$ nodes. Let $y_n[t]$ be the value of time series at time $\smash{t=0,1,\dots,T-1}$ observed  at node $\smash{1\leq n \leq N}$.} A $P$-th order nonlinear VAR model assuming  additive  functional dependencies can be formulated as\\
 \resizebox{1\linewidth}{!}{
	\begin{minipage}{1.00\linewidth}
	\begin{align} \label{eq:var}
	y_n[t] =   \sum_{\n = 1}^{N} \sum_{p = 1}^{P}a_{n,\n}^{\p} f_{n,n'}^{(p)}(y_{\n}[t-p])+u_n[t],
	\end{align}
	\end{minipage}
}
where $f_{n,n'}^{(p)}$ is the  function that encodes  the nonlinear causal influence of the $p$-lagged data at node $\n$ on the node $n$, \cB{$a_{n,\n}^{(p)}$ is the corresponding entry of the graph adjacency matrix, and $u_n[t]$ is the observation  noise.} Considering the model  \eqref{eq:var}, topology identification can be defined as the estimation of \cB{the functional dependencies $\cbr{a_{n,\n}^{\p}f_{n,n'}^{(p)}(.)}_{p=1}^P$} for $n=1,2,\dots,N$ from the observed time series $\cbr{y_n[t]}_{n=1}^N$.

We assume that the functions $f_{n,n'}^{(p)}$ in \eqref{eq:var} belong to a reproducing kernel Hilbert space (RKHS):\\
 \resizebox{1\linewidth}{!}{
	\begin{minipage}{1.15\linewidth}
\begin{align}\label{eq:rkhs}
\mathcal{H}_{n'}^\p:=\cbr{{f}_{n,n'}^{\p}~ |~ {f}_{n,n'}^{\p}\nbr{y}=\sum_{t=0}^{\infty}\beta^\p_{n,n',t}~\kappa^\p_{n'}\nbr{y,y_\n[t-p]}},
\end{align}
\end{minipage}
}
 \cB{where $\kappa^\p_\n: \mathbb{R}\times \mathbb{R} \rightarrow \mathbb{R}$ is the kernel associated with the Hilbert space. The kernel measures the similarity between data points $y$ and $y_\n[t-p]$. Referring to \eqref{eq:rkhs}, evaluation  of the functional ${f}_{n,n'}^{\p}$  at $y$ can be represented as the linear combination of the similarities between $y$ and the data points $\cbr{y_\n[t-p]}_{t=0}^{t=\infty}$, with weights $\beta^\p_{n,n',t}$. The inner product, \small{$\langle \kappa^\p_\n(y,x_1),\kappa^\p_\n(y,x_2)\rangle:=\sum_{t=0}^{\infty}\kappa^\p_\n(y[t],x_1)\kappa^\p_\n(y[t],x_2)$}, is defined in the Hilbert space using kernels with reproducible property $\langle\kappa^\p_\n(y,x_1),\kappa^\p_\n(y,x_2)\rangle=$ $\kappa^\p_\n(x_1,x_2)$. Such a Hilbert space with the reproducing kernels is termed as RKHS and the inner product described above induces a norm,  $\|{f}_{n,n'}^{\p} \|^2_{\mathcal{H}_{n'}^\p} =\sum_{t=0}^{\infty}\sum_{t'=0}^{\infty}\beta^\p_{n,n',t}~\beta^\p_{n,n',t'}~\kappa^\p_\n(y_n[t],y_n[t'])$.  We refer to \cite{Wahba1990} for further reading on RKHS.}
 
For a particular node $n$, the  estimates of $\cbr{{f}_{n,\n}^{\p}\in \mathcal{H}_{n'}^\p}_{\n,p}$ are obtained by solving the functional optimization problem:% \cG{ since we can decompose problem  in $n$: }
% \resizebox{1\linewidth}{!}{
% 	\begin{minipage}{1.00\linewidth}
 \begin{align} \label{eq: ls}
\cbr{\hat{{f}}_{n,\n}^\p}_{\n,p}=
\arg &\min_{\cbr{{f}_{n,\n}^{\p}\in \mathcal{H}_\n^\p}} \frac{1}{2}\sum_{\tau=P}^{T-1}\sbr{y_{n}[\tau]- \nonumber\\ &\sum_{\n = 1}^{N} \sum_{p = 1}^{P}a_{n,\n}^\p {f}_{n,\n}^\p(y_\n[\tau-p])}^2.
\end{align}
% \end{minipage}
% }
\cB{It is to be noted that in \eqref{eq: ls}, the functions $\{{f}_{n,\n}^{\p}\}$ belong to the RKHS defined in \eqref{eq:rkhs}, which is an infinite dimensional space. 
However, by resorting to the Representer Theorem} \cite{ageolk2000}, \cB{the solution of \eqref{eq: ls} can be written using a finite number of data samples:}  
\begin{align}\label{eqn:rep}
    \hat{f}_{n,n'}^{\p}\nbr{y_\n[\tau-p]}~~~~&\nonumber\\
    =\sum_{t=p}^{p+T-1}\beta^\p_{n,n',(t-p)}&\kappa^\p_{n'}\nbr{y_\n[\tau-p]),y_\n[t-p]}.
\end{align}
Notice that the number of coefficients required to express the function increases with the number of data samples. In the recent works \cite{only.s2018}, \cite{onlroh2021}, this problem is solved by using a time window to fix the number of data points, resulting in suboptimality. However, in this work in align with \cite{larjin2016} and \cite{ransjen2019}, we use RF approximation to tackle the dimensionality growth.  

\section{Random Feature approximation}
%\cR{To tackle curese of dimsnionality, this section reformulates the equation \eqref{eqn:rep} using random feature approximation. Random feature approximation is proposed based on the assumption that the kernel under consideration is shift-invariant.}
To invoke the RF approximation, we assume the kernel to be shift invariant, i.e., $\kappa^\p_{n'}\nbr{y_\n[\tau-p]),y_\n[t-p]}=\kappa^\p_{n'}\nbr{y_\n[\tau-p])-y_\n[t-p]}$. Bochner's theorem \cite{lecsal1959} states that every shift-invariant kernel can be represented as an inverse Fourier transform of a probability distribution. Hence the kernel evaluation can be expressed as
\begin{align}\label{eqn:bouchner}
  \kappa^\p_{n'}&\nbr{y_\n[\tau-p]),y_\n[t-p]}\nonumber\\&=\int \pi_{\kappa_{n'}^{(p)}}(v)~ e^{jv\nbr{y_\n[\tau-p]-y_\n[t-p]}}dv\nonumber\\
  &= \mathbb{{E}} _v[e^{jv\nbr{y_\n[\tau-p]-y_\n[t-p]}}],
\end{align}
where $\pi_{\kappa_{n'}^{(p)}}(v)$ is the probability density function which depends on type of the kernel, and $v$ is the random variable associated with it.
 If sufficient amount of iid samples $\{v_i\}_{i=1}^D$ are collected from the distribution $\pi_{\kappa_{n'}^{(p)}}(v)$, the real ensemble mean in \eqref{eqn:bouchner} can be expressed as a sample mean:
 \begin{align}\label{eqn:sample}
 \hat{\kappa}^\p_{n'}&\nbr{y_\n[\tau-p]),y_\n[t-p]}=\nonumber\\\frac{1}{D}&\sum_{i=1}^De^{jv_i\nbr{y_\n[\tau-p])-y_\n[t-p]}},
 \end{align}
 irrespective of the distribution $\pi_{\kappa_{n'}^{(p)}}(v)$. Note that the unbiased estimate of kernel evaluation in \eqref{eqn:sample} involves a summation of fixed $D$ number of terms. In general, computing the probability distribution corresponding to a kernel is a difficult task. In this work the kernel under consideration is Gaussian; for a Gaussian kernel
 $k_{\sigma}$ with variance $\sigma^2$, it is well known that the Fourier transform is  a Gaussian with variance $\sigma^{-2}$. Considering the real part of \eqref{eqn:sample}, which is also an unbiased estimator, \eqref{eqn:bouchner} can be approximated as
 \small
 \begin{align}
  &\hat{\kappa}^\p_{n'}\nbr{y_\n[\tau-p],y_\n[t-p]}=\boldsymbol{z_v}\nbr{y_\n[\tau-p]}^\top\boldsymbol{z_v}\nbr{y_\n[t-p]}, \label{eqn:rfaprox} \shortintertext{where} 
  &\boldsymbol{z_v}(x)=\frac{1}{\sqrt{D}}[\sin
  v_1x,\dots,\sin v_Dx,\cos v_1x,\dots,\cos v_Dx]^\top. \label{eqn:feature}
\end{align}
\normalsize
 Subsisting \eqref{eqn:rfaprox} in \eqref{eqn:rep}, we obtain a fixed dimension ($2D$ terms) approximation of the function $\hat{f}_{n,n'}^{\p}$:
 
\begin{align}\label{eqn:newrep}
    &\hat{\vphantom{\rule{2pt}{6.0pt}}\smash{\hat{f}}}_{n,n'}^{\p}\nbr{y_\n[\tau-p])}\nonumber\\&=\sum_{t=p}^{{p+T-1}}\beta^\p_{n,n',(t-p)}\boldsymbol{z_v}\nbr{y_\n[\tau-p]}^\top\boldsymbol{z_v}\nbr{y_\n[t-p]}\nonumber\\&={\boldsymbol{\theta}_{n,n'}^{(p)}}^\top\boldsymbol{z_v}\nbr{y_\n[\tau-p]},
\end{align}
where ${\boldsymbol{\theta}_{n,n'}^{(p)}}^\top=\sum_{t=p}^{{p+T-1}}\beta^\p_{n,n',(t-p)}\boldsymbol{z_v}\nbr{y_\n[\tau-p]}^\top=[\theta_{n,\n,1}^\p, \dots,\theta_{n,\n,2D}^\p  ] \in \mathbb{R}^{2D}$.  For the sake of clarity, in the succeeding steps, we define the following notation:%express, ${\boldsymbol{\alpha}_{n,n'}^{(p)}}$ and $\boldsymbol{z_v}\nbr{y_\n[\tau-p]}$ as follows:
\begin{align}
{\boldsymbol{\alpha}_{n,n'}^{(p)}}&=[\alpha_{n,\n,1}^\p, \dots,\alpha_{n,\n,2D}^\p  ]^{\top} \in \mathbb{R}^{2D},\\
\boldsymbol{z_v}\nbr{y_\n[\tau-p]}&=[ z^\p_{n',1}\nbr{\tau}, \dots z^\p_{n',2D}\nbr{\tau}]^{\top}\in \mathbb{R}^{2D},
\label{eqn:zthetavec}
\end{align}
where $\alpha_{n,\n,d}^\p=\theta_{n,\n,d}^\p a_{n,\n}^\p$. The functional optimization \eqref{eq: ls} is reformulated as a parametric optimization problem using \eqref{eqn:newrep}:
 \begin{align} \label{eq: ParOpt}
\cbr{\widehat{\alpha}_{n,\n,d}^\p}_{\n,p,d}=
\arg &\min_{\cbr{\alpha_{n,\n,d}^\p}} \mathcal{L}^n\nbr{\alpha_{n,\n,d}^\p},
\end{align}
where
 \begin{align} \label{eq: ls2}
\mathcal{L}^n\nbr{\alpha_{n,\n,d}^\p}&:=\sum_{\tau=P}^{T-1}\frac{1}{2}\sbr{y_{n}[\tau]- \sum_{\n = 1}^{N} \sum_{p = 1}^{P}\sum_{d=1}^{2D}\alpha^\p_{n,n',d}~ z^\p_{n',d}\nbr{\tau}}^2.
\end{align}
For convenience,  optimization parameters  $\cbr{{\alpha}^\p_{n,n',d}}$ and $\cbr{z^\p_{n',d}\nbr{\tau}}$ are stacked in the  lexicographic order of the indices $p$, $\n$, and $d$ to obtain the vectors $\bs {\alpha}_n\in \mathbb{R}^{2PND}$ and $\boldsymbol{z}_\tau\in \mathbb{R}^{2PND}$, respectively, and \eqref{eq: ParOpt} is rewritten as
\begin{align} 
\widehat{\bs{\alpha}}_n=
\arg &\min_{\bs\alpha_{n}}  \mathcal{L}^n\nbr{\bs{\alpha}_{n}}, \label{eq: ParOpt2}\\
\text{where }~
\mathcal{L}^n(\bs{\alpha}_n)&=\frac{1}{2}\sum_{\tau=P}^{T-1}\sbr{y_{n}[\tau]- \bs{\alpha}_{n}^\top\bs{z}_\tau}^2 \label{eq: ls4}
\end{align}
Now, in order to avoid overfitting,  we propose a regularized optimization framework:
\begin{align} \label{eq: ParOpt2}
\widehat{\bs{\alpha}}_n=
\arg &\min_{\bs\alpha_{n}}  \mathcal{L}^n\nbr{\bs{\alpha}_{n}}+\lambda \sum_{\n = 1}^{N} \sum_{p = 1}^{P}\| \bs{\alpha}_{n,n'}^\p \|_2,
\end{align}
where $\lambda\geq0$ is the regularization parameter and $\bs{\alpha}_{n,n'}^\p =(\alpha_{n,n',1}^\p,\alpha_{n,n',2}^\p,\dots,\alpha_{n,n',2D}^\p)\in \mathbb{R}^{2D}.$
The second term in \eqref{eq: ParOpt2} is a \textit{group-lasso} regularizer, which promote a \textit{group-sparse structure} in $\bs{\alpha}_{n,n'}^\p$,  supported by the assumption that most of the real world dependencies are sparse in nature.

However, notice that the batch formulation in \eqref{eq: ParOpt2} has some significant limitations: \textbf{i)} requirement of complete batch of data points before estimation, \textbf{ii)} inability to track time varying topologies, and \textbf{iii)} explosive computational complexity when $T$ is large even if RF approximation is used. To mitigate these problems, we adopt an online optimization strategy, which is explained in the following section.   
\section{Online topology estimation}
    In this case, we replace the batch loss function $\mathcal{L}^n(\bs\alpha_n)$ in \eqref{eq: ParOpt2} with the stochastic (instantaneous) loss function $l_t^n(\bs\alpha_n)=\frac{1}{2}[y_{n}[t]- \bs{\alpha}_{n}^\top\bs{z}_t]^2$:
\begin{align} \label{eq: parOpt3}
\widehat{\bs{\alpha}}_n=
\arg &\min_{\bs\alpha_{n}}  l_t^n\nbr{\bs{\alpha}_{n}}+\lambda \sum_{\n = 1}^{N} \sum_{p = 1}^{P}\| \bs{\alpha}_{n,n'}^\p \|_2.
\end{align}
Notice that the   sparsity promoting group lasso regularizer  is non-differentiable. The use of online subgradient descent (OSGD) is not advisable in this situation as it  linearizes the entire objective function and fails to provide sparse iterates. To avoid this limitation of OSGD, we use the composite objective mirror descent (COMID) \cite{Comjoh2010} algorithm which resembles the nature of proximal methods, hence improving convergence. The online COMID update can be written as 
    \begin{align}
    {\bs{\alpha}}_n&[t+1]=\arg \min_{{\bs{\alpha} }_n} J_t^{(n)}({\bs{\alpha}}_n),\label{eqn:comid_update}\\
    &\text{where}~ J_t^{(n)}({\bs{\alpha}}_n) \triangleq \nabla \ell_{t}^n({\bs{{\alpha}}}_n[t])^\top\nbr{{\bs{\alpha}}_n -{\bs{{\alpha}}}_n[t]}\nonumber\\ &+\frac{1}{2\gamma_t}\| {\bs{\alpha}}_n -{\bs{{\alpha}}}_n[t]\|_2^2+\lambda \sum_{\n = 1}^{N} \sum_{p = 1}^{P}\| \bs{\alpha}_{n,n'}^\p \|_2. \label{eqn:comid_risk}
\end{align}
In \eqref{eqn:comid_risk} ${\bs {{\alpha}}}_n[t]\in \mathbb{R}^{2PND}$ denotes the estimate of  $\bs \alpha_n$ at time $t$. The first term in equation \eqref{eqn:comid_risk} is the gradient of the loss function $l_t^n(\bs\alpha_n)$, the second and third term are Bergman divergence and sparsity promoting regularizer respectively. The Bregman divergence is included to improve the stability of algorithm from adversaries by constraining ${\bs \alpha}_n[t+1]$ to be close to ${\bs{ {\alpha}}}_n[t]$. The Bregman divergence $B({\bs{\alpha}}_n ,{\bs{{\alpha}}}_n[t])=\frac{1}{2}\| {\bs{\alpha}}_n -{\bs{{\alpha}}}_n[t]\|_2^2$ chosen in such a way that the COMID update has a closed form solution \cite{bregut2011} and $\gamma_t$ is the corresponding step size.
 The gradient in \eqref{eqn:comid_risk} is evaluated as
\begin{align}
{\bf v}_n[t]&:=\nabla \ell_{t}^n({\bs{\alpha}}_n[t])= \boldsymbol{z}_t\nbr{\bs \alpha_{n}^\top\bs{z}_t-y_{n}[t]}
% {\bf v}_n[t]&={\bf \boldsymbol{\kappa}}_\tau\nbr{\bs \alpha_{n}^\top\bs\kappa_\tau-y_{n}[\tau]}
\label{eqn:grad}
\end{align}
 Expanding the objective function in \eqref{eqn:comid_risk} and  omitting the constants leads to the following formulation:

 \resizebox{0.9\linewidth}{!}{
	\begin{minipage}{\linewidth}
\begin{align} 
\hspace{-1cm}
    J_t^{(n)}({\bs{\alpha}}_n) &\propto \frac{{\bs{\alpha}}_n^\top{\bs{\alpha}}_n}{2\gamma_t}+{\bs{\alpha}}_n^\top\nbr{{\bf v}_n[t]-\frac{1}{\gamma_t}{\bs{ \alpha}}_n[t]} +\lambda \sum_{\n = 1}^{N} \sum_{p = 1}^{P}\| \bs{\alpha}_{n,n'}^{(p)} \|_2\nonumber
    \end{align}
\end{minipage}
}
\begin{align}
      =\sum_{\n=1}^N \sum_{p = 1}^{P}&\sbr{\frac{{\bs \alpha}_{n,\n}^{{\p}^\top}{\bs \alpha}_{n,\n}^{\p}}{2\gamma_t}+{\bs \alpha}_{n,\n}^{{\p}^\top}\nbr{{\bf v}_{n,\n}^{\p}[t]-\frac{1}{\gamma_t}{\bs {{\alpha}}}_{n,\n}^{\p}[t]}~~~~~\nonumber\\
      &+\lambda  \| \bs \alpha_{n,n'}^{(p)} \|_2}.\label{eqn:comid_risk3}
\end{align}
A closed form solution for \eqref{eqn:comid_update} using \eqref{eqn:comid_risk3} is obtained via the multidimensional shrinkage-thresholding operator \cite{amupui2009}:
\begin{align}
    {\bs \alpha}_{n,\n}^{(p)}[t+1]&=\nbr{{\bs {{\alpha}}}_{n,\n}^{(p)}[t]-\gamma_t{\bf v}_{n,\n}^{(p)}[t]}\times \nonumber 
    \\ &\sbr{1-\frac{\gamma_t\lambda}{\|{\bs {{\alpha}}}_{n,\n}^{(p)}[t]-\gamma_t{\bf v}_{n,\n}^{(p)}[t]\|_2}}_+,
    \label{eqn:sol}
\end{align}
 where $[x]_+=\max\cbr{0,x}$. The first term ${\bs {{\alpha}}}_{n,\n}^{(p)}[t]-\gamma_t{\bf v}_{n,\n}^{(p)}[t]$ in \eqref{eqn:sol} forces the  stochastic gradient update of  $\bs {\alpha}_{n,\n}^{(p)}$ in  a way to descend  instantaneous loss function $l_t^n(\bs\alpha_n)$ and the second term in \eqref{eqn:sol} enforces group sparsity of $\bs {\alpha}_{n,\n}^{(p)}$. Note that the close form solution \eqref{eqn:sol} is separable in $n'$ and $p$.
 
The proposed algorithm, termed as \textit{Random Feature based Nonlinear Topology Identification via Sparse Online learning} (RF-NLTISO), is summarized in Algorithm 1. %\cref{alg:RF-NLTISO}.
\begin{algorithm}[h!]
 \label{alg:RF-NLTISO}
\SetAlgoLined
\KwResult{$\boldsymbol{\alpha}_{n,n'}^\p,  for~ n,n'=1,..,N$ and~$p=1,..,P$ }
 \textbf{Store} $\{\boldsymbol{y}_n[t]\}_{t=1}^P$,\\
 \textbf{Initialize} $\lambda$, $\gamma$, $D$ (heuristically chosen) and  kernel parameters depending on the type of the kernel. \\
 \For{$t=P,P+1,\dots$}{
  Get data samples ${y}_n[t],~\forall n$ and compute $\boldsymbol{z}_\tau$
  \\\For{$n=1,\dots,N$}{
  compute $\bm{v}_n[t]$ using \eqref{eqn:grad}
 \\ \For{$\n=1,\dots, N$}{
  compute $\boldsymbol{\alpha}_{n,n'}^\p[t+1] $ using \eqref{eqn:sol} }
  }
 }
 \caption{RF-NLTISO Algorithm}
\end{algorithm}
\section{Experiments}
We compare the performance of the proposed algorithm, RF-NLTISO, with the the state-of-the-art online topology estimation algorithms. Experiments shown in this section are conducted using 1) synthetic datasets with topologies having different transition patterns and 2) real datasets collected from Lundin's offshore oil and Gas platform. For the performance comparison, we choose TIRSO \cite{onlzam2019} and NL-TISO \cite{onlroh2021} algorithms, which are the state-of-the-art counterparts of RF-NLTISO, to the best of our knowledge. TIRSO is developed based on a linear VAR model assumption, whereas NL-TISO, a kernel-based topology estimation algorithm, is developed for nonlinear VAR models. Although a kernel-based functional stochastic gradient based algorithm \cite{only.s2018} is also available, its performance has been shown to be inferior compared to NL-TISO \cite{onlroh2021}. %\textcolor{red}{It is observed that the proposed algorithm outperforms TIRSO and NL-TISO and it has a remarkable improvements in computational complexity compared to NL-TISO \cite{onlroh2021}.}
\subsection{Experiments using Synthetic Data}
\subsubsection{Topology with switching edges}
\label{Tid}
We generate a multi-variate time series using nonlinear VAR model \eqref{eq:var} with $N=5,P=2$. An initial random graph with edge probability of $0.1$ is generated and the graph adjacency coefficients $a_{n,\n}^{\p}$ are drawn from a Uniform distribution $\mathcal{U}(0,\,1)$. After every $1000$ samples, one of the active (non-zero) edge disappears and another one appears randomly, which brings an abrupt change in the graph topology. The nonlinearity in \eqref{eq:var} is introduced using a Gaussian kernel with variance $0.01$ and the kernel coefficients are chosen  randomly from a zero mean Gaussian distribution with variance $30$. Note that the initial $P$ data samples are generated  randomly and rest of the data is generated using the model \eqref{eq:var}.
\begin{figure*}[h!]
	\centering
	\begin{subfigure}{0.3\textwidth}
	\centerline{\includegraphics[scale=0.1,trim={0cm 2cm 0cm 0cm}, clip]{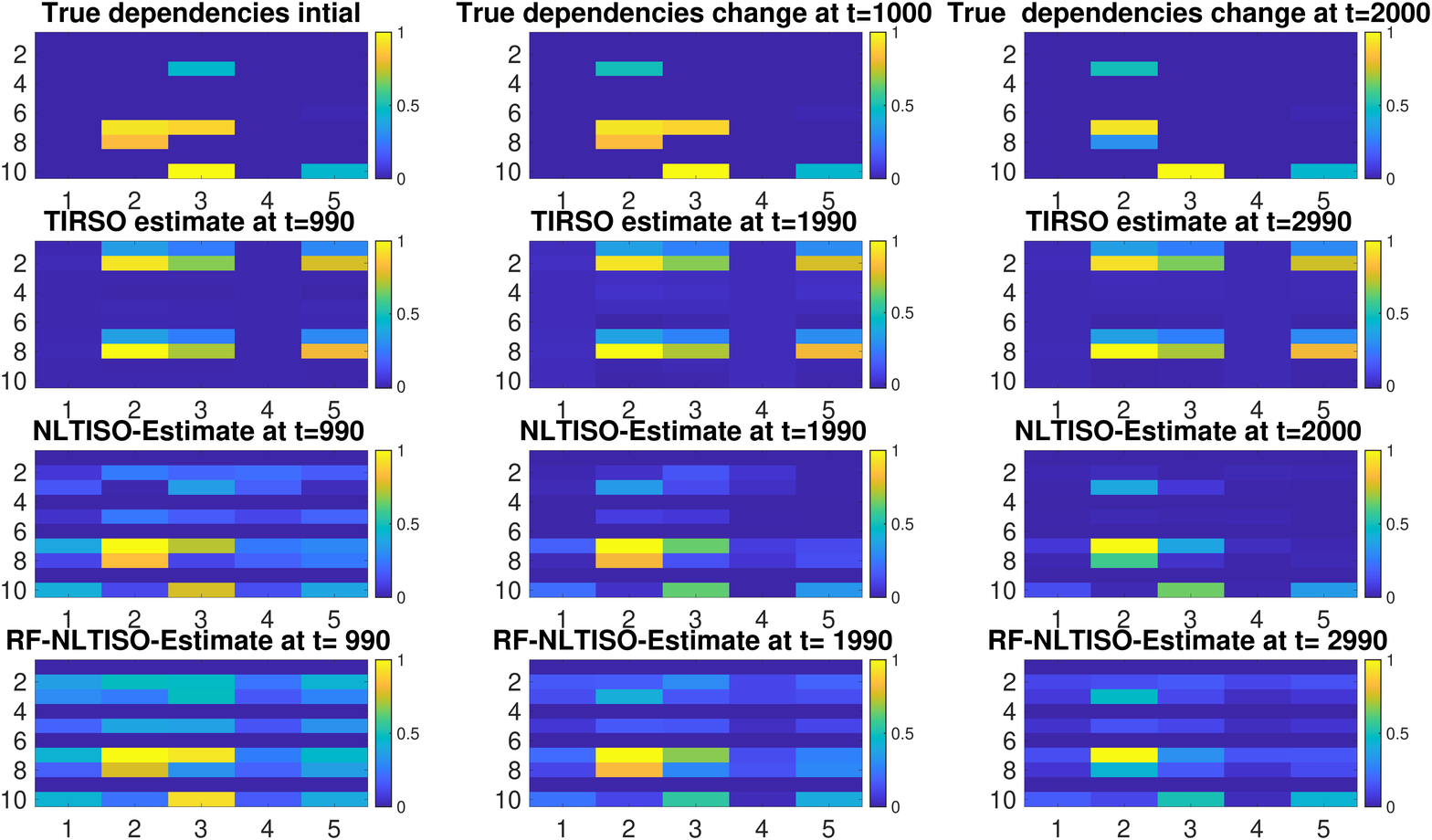}}
		\caption{Causal dependencies  estimated using different algorithms compared with the  true dependency.}
		\label{topid}
	\end{subfigure}
	\hfill
	\begin{subfigure}{0.12\textwidth}
    	\centerline{\includegraphics[scale=0.21,trim={0cm 0.1cm 1cm 0cm}, clip]{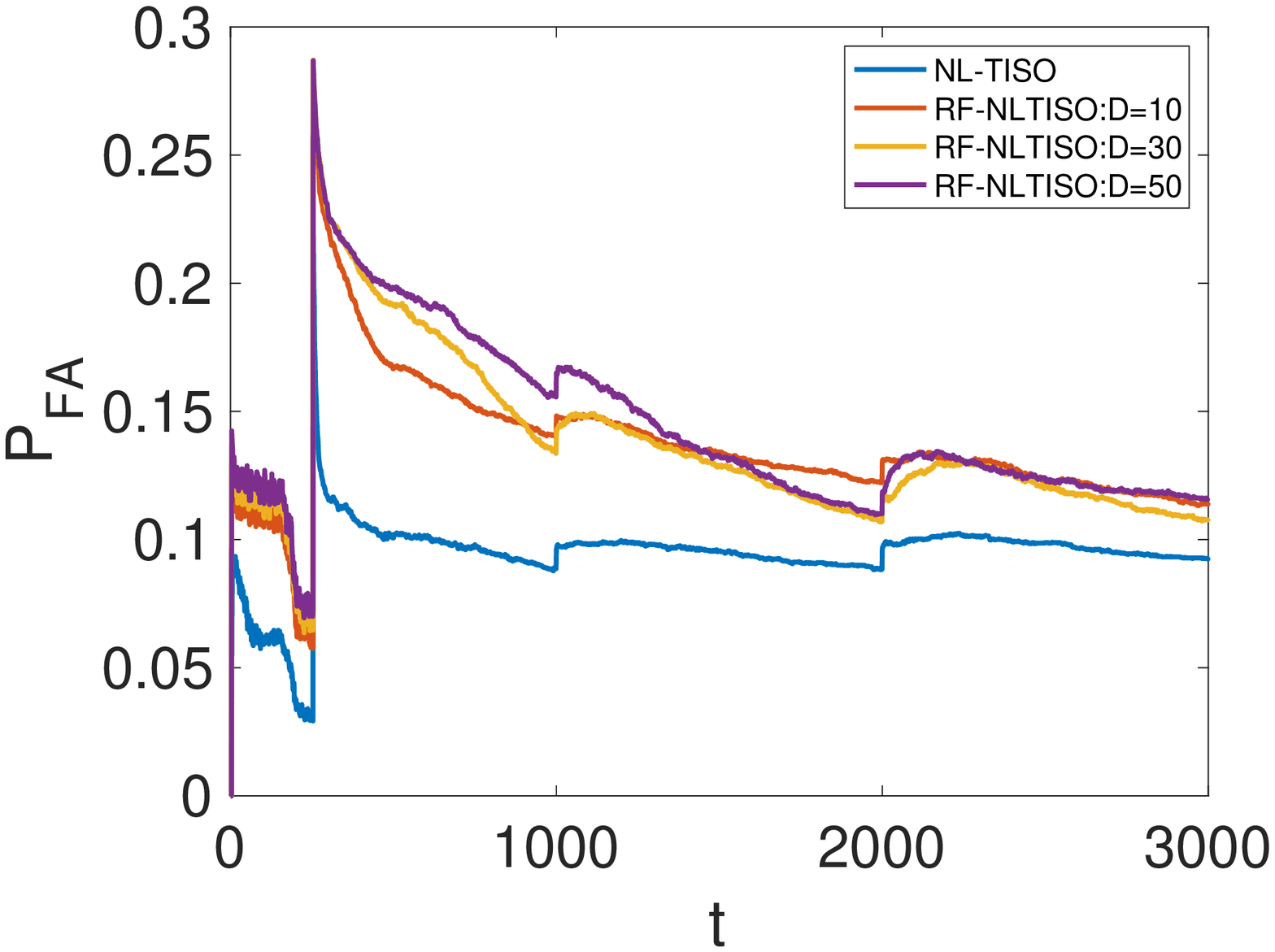}}
        \caption{Probability of False Alarm ($P_{FA}$). }
        \label{pfa}
	\end{subfigure}
	\hfill
	\begin{subfigure}{0.12\textwidth}
    	\centerline{\includegraphics[scale=0.21,trim={0cm 0.1cm 1cm 0cm}, clip]{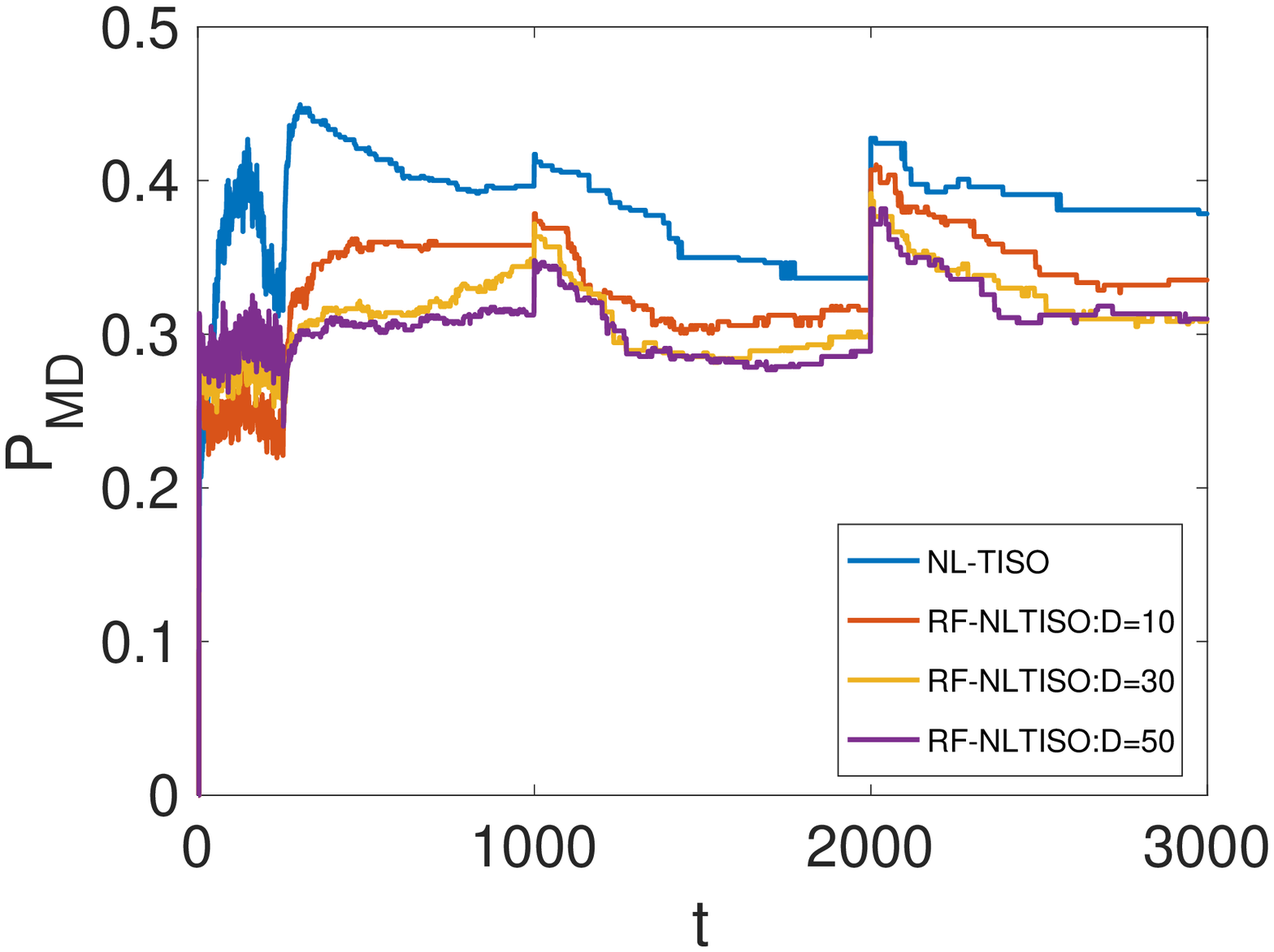}}
        \caption{Probability of Miss Detection ($P_{MD}$).}
        \label{pmd}
	\end{subfigure}
	\hfill
	\begin{subfigure}{0.12\textwidth}
        \centerline{\includegraphics[scale=0.21,trim={0cm 0cm 1cm 0.8cm}, clip]{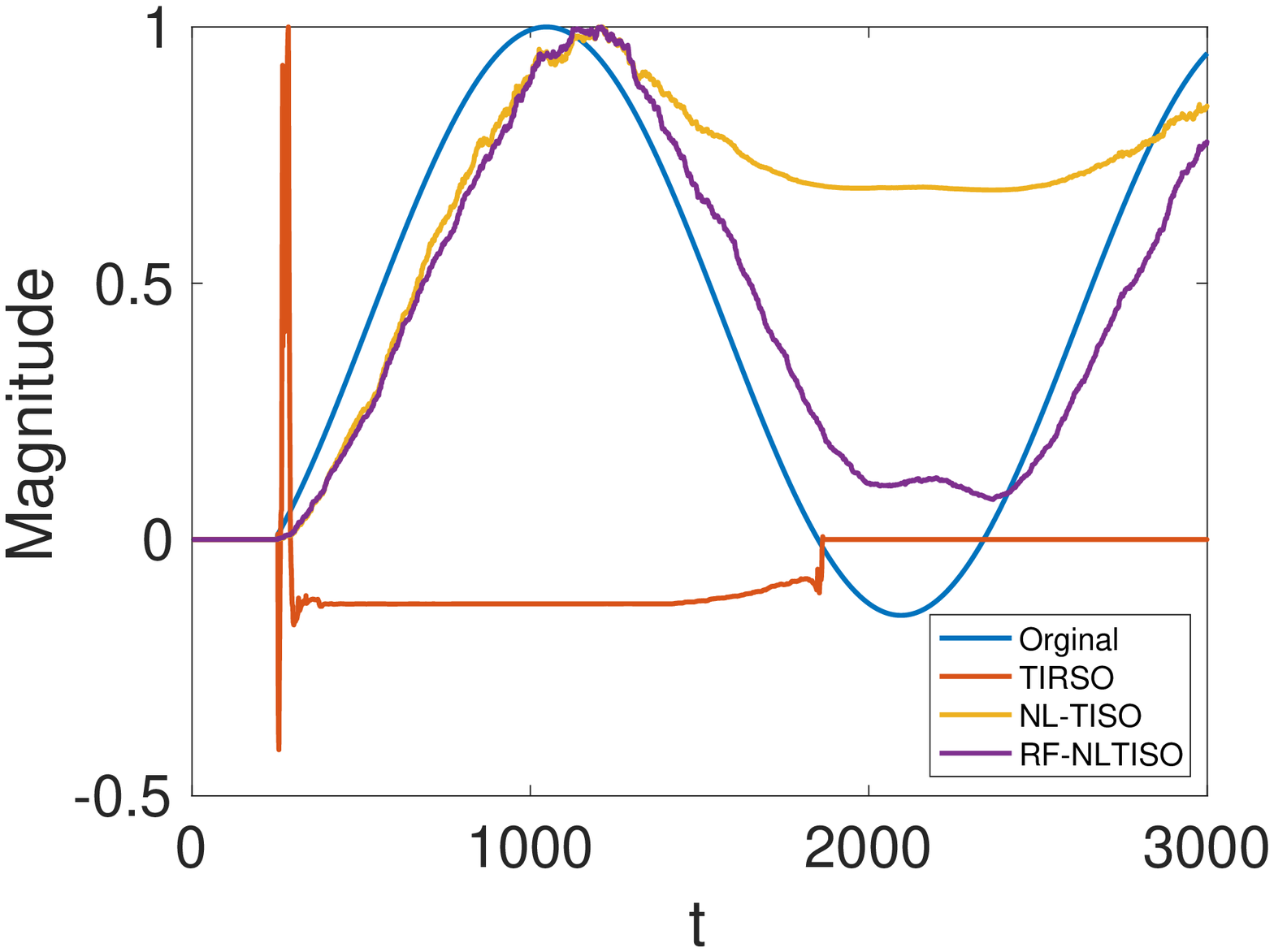}}
        \caption{Slowly varying dependency.}
        \label{slow}
	\end{subfigure}
	\caption{ }
	\label{subfig_1}
\end{figure*}

\cB{The coefficients $\cbr{{\bs \alpha}_{n,\n}^{(p)}[t]}$ are estimated using the proposed RF-NLTISO algorithm with a Gaussian kernel having variance $0.1$ and number of random features $D=50$. The hyper-parameters  $\lambda$ and $\gamma$ are heuristically chosen as $0.1$ and $1000$, respectively. We compute the $\ell_2$ norms $\widehat{b}_{n,n'}^\p[t]=\|\bs{\alpha}_{n,n'}^\p[t]\|_2$ and  arrange them in a matrix form similar to the graph adjacency matrix to visualize the causal dependencies. A similar strategy is adopted for the NL-TISO and the TIRSO algorithms. The normalized version of true and the estimated dependencies at various time samples are  shown in \cref{topid}, where in  each subplot, the $5\times5$ dependency matrices corresponding to $p=1 \rm{~and~}2$ are concatenated, resulting in a $10\times 5$ size matrix. We normalized the coefficients by dividing each coefficients with highest value of coefficient in a pseudo adjacency matrix.} From the \cref{topid}, it is clear that RF-NLTISO is able to perform equal or better compared to NL-TISO algorithm and clearly outperforms TIRSO. 

Next we conduct the same experiments using RF-NLTISO with different numbers of random feature  ($D\in\cbr{10,30,50}$). These experiments are repeated 1000 times to find probability of miss detection ($P_\text{MD}$) and false alarm ($P_\text{FA}$), which we define as
\small
\begin{align}
P_\text{MD}[t] \triangleq \frac{\sum _{n\neq \n}\sum _{p=1}^P \mathbb{E}\left[ \mathbb {1}\lbrace \| {\widehat{\boldsymbol b}}^\p_{n,n^{\prime }}[t]\| _2 < \delta \rbrace \mathbb {1}\lbrace \left\Vert \boldsymbol \alpha_{n,n^{\prime }}\right\Vert _2 \ge \delta \rbrace \right] }{\sum _{n\neq\n}\sum _{p=1}^P \mathbb{E}\left[ \mathbb {1}\lbrace \left\Vert \boldsymbol \alpha_{n,n^{\prime }}\right\Vert _2 \ge \delta \rbrace \right]},\nonumber \\
P_\text{FA}[t] \triangleq \frac{\sum _{n\neq \n}\sum _{p=1}^P \mathbb{E}\left[ \mathbb {1}\lbrace \| {\widehat{\boldsymbol b}}^\p_{n,n^{\prime }}[t]\| _2 > \delta \rbrace \mathbb {1}\lbrace \left\Vert \boldsymbol \alpha_{n,n^{\prime }}\right\Vert _2 \le \delta \rbrace \right] }{\sum _{n\neq\n}\sum _{p=1}^P \mathbb{E}\left[ \mathbb {1}\lbrace \left\Vert \boldsymbol \alpha_{n,n^{\prime }}\right\Vert _2 \le \delta \rbrace \right]}.
\end{align}
\normalsize
From \cref{pfa,pmd}, it is observed that for the given choices of $D$, $P_{MD}$ is better for RF-NLTISO compared to NL-TISO; however, NL-TISO is performing better in terms of $P_{FA}$. Both the figures show an overshoot at the topology-switching time instances. It is also observed that for the proposed algorithm, $P_{MD}$ decreases with $D$, whereas $P_{FA}$ increases with $D$, which in turn suggests a tuning for $D$ for an effective trade-off between $P_{FA}$ and  $P_{MD}$
\subsubsection{Slowly varying topology}
\label{srec}
We compare the performance of RF-NLTISO with the state-of-the-art algorithms using a slowly varying graph topology. The same experiment setup as discussed in \cref{Tid} is adopted with the following more slowly  time varying topology: 
\begin{align}\label{eq:sine1}
    a^\p_{n,n'}[t+1]=a^\p_{n,n'}[t]+0.01\sin(0.03*t)
\end{align}
The normalized values of one of the active edges is plotted in \cref{slow}. The figure also shows the normalized values of the corresponding estimated coefficients ($\widehat{b}_{n,n'}^\p[t]$ for NL-TISO and RF-NLTISO and $\widehat{a}_{n,n'}^\p[t]$ for TIRSO ). From the figure, it can be observed that the RF-NLTISO estimates are closer to the true value compared to the estimates from the other two algorithms. In this example, the quality of TIRSO estimates lags considerably behind the kernel-based algorithms due to the fact that the underlying VAR model is nonlinear.

\subsection{Experiments using Real Data}
\label{real}
\cB{This section is dedicated to  experiments using real data collected from Lundin’s offshore oil and gas (O\&G) platform Edvard-Grieg\footnote{https://www.lundin-energy.com/}. We have a multi-variate time series  with  $24$ nodes; and the  nodes corresponds to various  temperature (T), pressure (P), or oil-level (L) sensors. The sensors are placed in the separators of  decantation tanks that separate oil, gas, and water. The time series are obtained by  uniformly sampling the sensor readings with a sampling rate of $5s$. We assume that hidden logic  dependencies  are present in the network due to various physical  connections and various control actuators. The data obtained from the network is normalized by making it a   zero mean unit variance signal,  before applying the algorithm.}
\begin{figure*}[h!]
	\centering
	\begin{subfigure}{0.3\textwidth}
	\centerline{\includegraphics[scale=0.3]{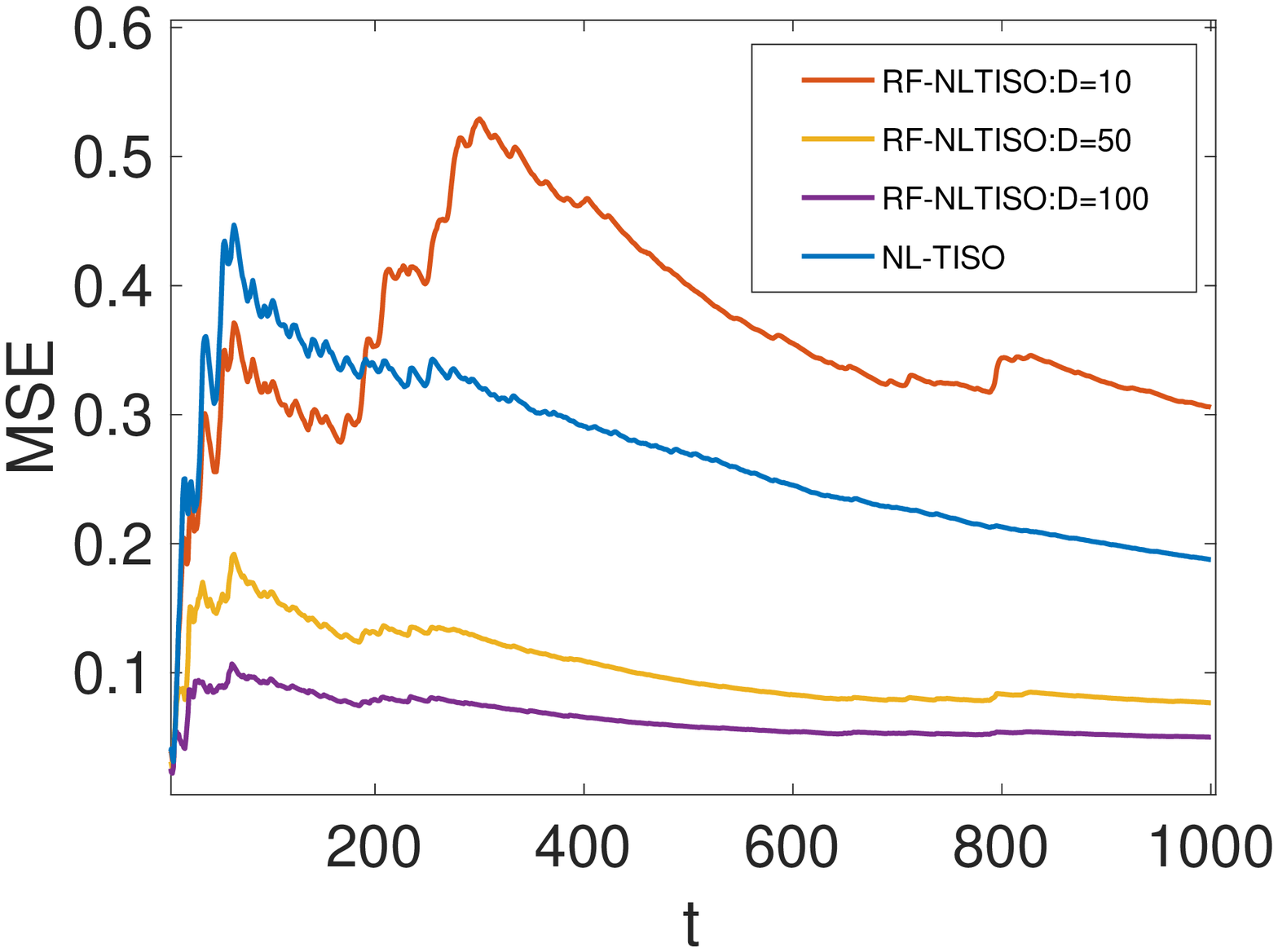}}
    \caption{MSE comparison of NL-TISO with RF-NLTISO.}
    \label{realerror}
	\end{subfigure}
	\hfill
	\begin{subfigure}{0.3\textwidth}
	\centerline{\includegraphics[scale=0.3]{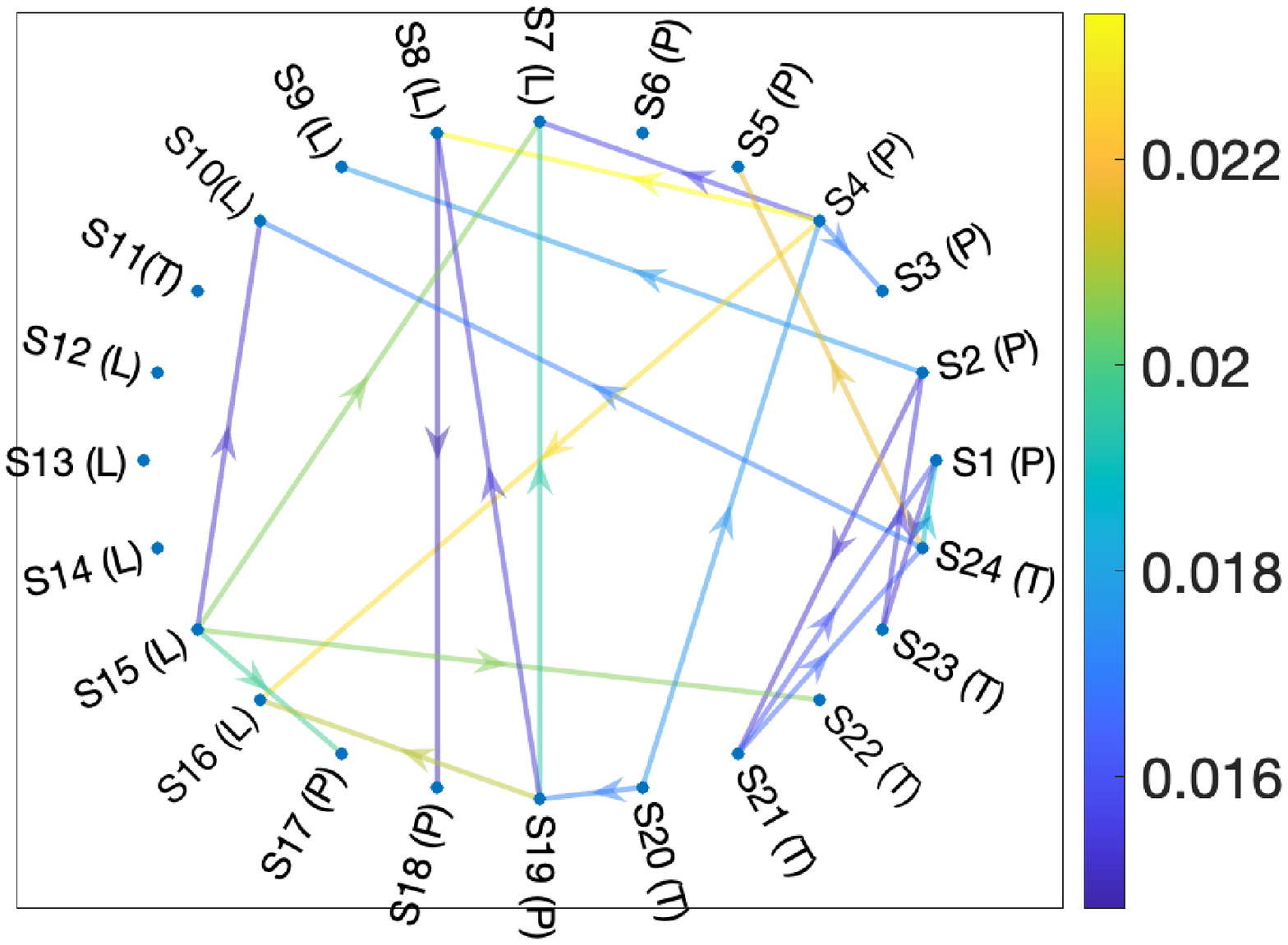}}
    \caption{Causality graph in oil and gas plant estimated by RF-NLTISO. P, T, L represent pressure, temperature, and oil level sensors, respectively.}
    \label{realtop}
	\end{subfigure}
	\hfill
	\begin{subfigure}{0.3\textwidth}
	\centerline{\includegraphics[scale=0.3]{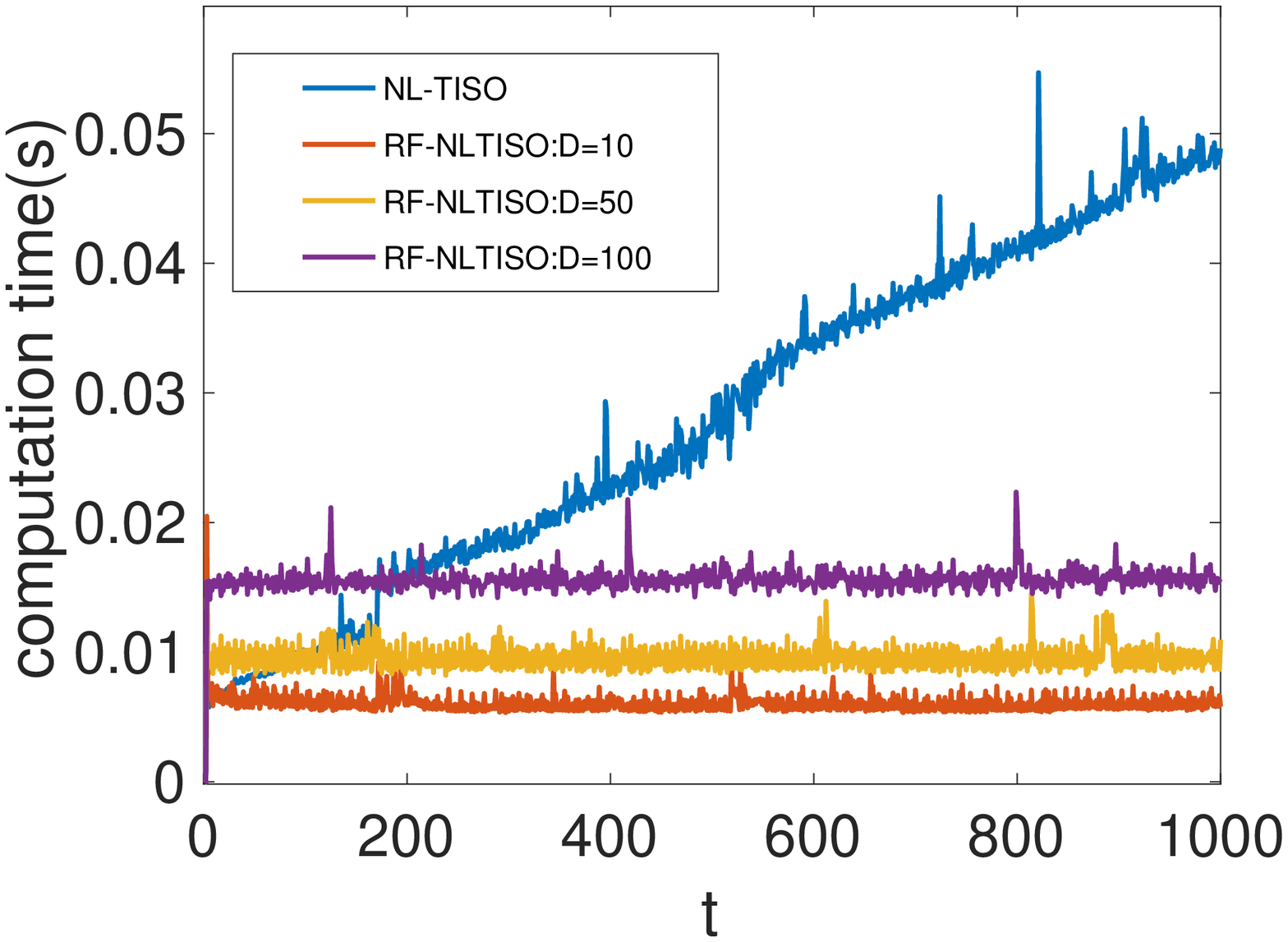}}
    \caption{Comparison of computation time of kernel-based algorithms}
    \label{comp}
	\end{subfigure}
	\hfill
	\caption{ }
	\label{subfig_2}
\end{figure*}
The causal dependencies are learned using RF-NLTISO with $D=10,50,100$ and a Gaussian kernel having a variance of $0.1$ and with hyper parameter values $\lambda=0.1$ and $\gamma=10$.
The signal is reconstructed using the estimated dependencies.
\Cref{realerror} shows the mean squared error ($MSE$), defined as $MSE(t)=\mathbb{E}((y_n(t)-\hat{y}_n(t))^2)$ for a particular sensor $n=8$, of RF-NLTISO estimates in comparison with other algorithms.  We observe that the RF-NLTISO estimates with random feature number $D\geq50$ show better $MSE$ performance compare to NL-TISO. The causality graph estimated by RF-NLTISO is shown in \cref{realtop}. %\cG{Where mse is defined as $MSE(t)=E((y_n(t)-\hat{y}_n(t))^2)$ for a particular sensor $n$}

One of the main attractiveness of RF-NLTISO is that even though it is a kernel-based algorithm, it has a fixed computational complexity throughout the online iterations. To demonstrate this, in \cref{comp}, we plot the computation time required to estimate the coefficients at each time instant by NL-TISO and RF-NLTISO with different values of $D$. The experiment is conducted in a machine with processor 2.4 GHz 8-core Intel Core i9 and 16GB 2667 MHz DDR4 RAM. \Cref{comp} shows that the computation time of NL-TISO increases considerably with time but that of RF-NLTISO remains more or less constant for a particular value of $D$.
 \section{Conclusion}
We propose a kernel-based online topology identification method for interconnected networks of time-series with additive nonlinear dependencies. In this work, the curse of dimensionality associated with kernel representation is tackled using random feature approximation. Assuming that the real-world dependencies are sparse, we use composite objective mirror decent update to estimate the online sparse causality graph. The effectiveness of the proposed algorithm is illustrated through experiments conducted on synthetic and real data, which shows that the algorithm outperforms the state-of-the-art competitors. We devote the convergence and stability analysis of the proposed algorithm to our future work.

%as the multiple signals we are dealing with might have different frequencies. Moreover, the frequencies of a particular signal itself can vary across time.

\bibliographystyle{IEEEbib}
\bibliography{strings,refs}

\end{document}